\documentclass[conference]{IEEEtran}
\IEEEoverridecommandlockouts
\usepackage{cite}
\usepackage{amsmath,amssymb,amsfonts}
\usepackage{algorithmic}
\usepackage{graphicx}
\usepackage{textcomp}
\usepackage{subcaption}
\usepackage{multirow}
\usepackage{xcolor}
\usepackage{flushend}
\def\BibTeX{{\rm B\kern-.05em{\sc i\kern-.025em b}\kern-.08em
    T\kern-.1667em\lower.7ex\hbox{E}\kern-.125emX}}
\begin{document}

\title{Salient Object Detection with Convex Hull Overlap}

\author{\IEEEauthorblockN{1\textsuperscript{st} Yongqing Liang}
\IEEEauthorblockA{\textit{School of Electrical Engineering and Computer Science} \\
\textit{Louisiana State University}\\
Baton Rouge, the United States of America\\
yqliang@cct.lsu.edu}
}

\maketitle

\begin{abstract}
Salient object detection plays an important part in a vision system to detect important regions. Convolutional neural network (CNN) based methods directly train their models with large-scale datasets, but what is the crucial feature for saliency is still a problem. In this paper, we establish a novel bottom-up feature named convex hull overlap (CHO), combining with appearance contrast features, to detect salient objects. CHO feature is a kind of enhanced Gestalt cue. Psychologists believe that surroundedness reflects objects overlap relationship. An object which is on the top of the others is attractive. Our method significantly differs from other earlier works in (1) We set up a hand-crafted feature to detect salient object that our model does not need to be trained by large-scale datasets; (2) Previous works only focus on appearance features, while our CHO feature makes up the gap between the spatial object covering and the object's saliency. Our experiments on a large number of public datasets have obtained very positive results.
\end{abstract}

\begin{IEEEkeywords}
    salient object detection, convex hull overlap, saliency map, Gestalt convexity, hand-crafted feature
\end{IEEEkeywords}

\section{Introduction}

Usually, people can detect visually distinctive noticeable foregrounds in the scene (that is, salient objects or regions) effortlessly and rapidly. This capability has long been studied by cognitive scientists and attracted a lot of research interests for salient region detection in the field of computer vision. Previous studies on salient region detection focus on two main strategies. The first one is called top-down or task-driven. It is determined by cognitive phenomena like knowledge, expectations, reward and current goals. The other one is called bottom-up. The selection of salient objects depends on low-level features (e.g., color, texture, geometry, overlap, etc.) of images. At present, most related research work has done well on the bottom-up-based pattern \cite{achanta2009frequency} \cite{cheng2015global} \cite{hou2007saliency} \cite{lu2011salient} \cite{zhai2006visual}. Such a pattern generally utilizes the regular features to exhibit the visual properties of image contents, which is effective for salient region detection but involves the potential need of the in-depth valuable feature discovery for more accurate detection. Thus, the salient region detection with more comprehensive feature information is becoming more popular to alleviate the above problems, which has been proved to be successful in more recent related work.

Convolutional neural networks (CNNs) have showed the impressive results in many image and video processing tasks. In recent years, the Fully Convolutional Neural Network (FCN) raised the bar in performance for dense predictions. FCN-based methods are able to extract more representative and complex features than hand crafted features \cite{lee2016deep} \cite{li2015visual} \cite{li2016deepsaliency} \cite{wang2015deep} \cite{zhao2015saliency}. However, FCN-based methods need to be trained by large pixel-level segmentation datasets. It is difficult to find out which factor that influences the object's saliency from CNN-based features. In this paper, we propose a hand crafted feature to explain how the objects' overlap positions affects people's attention.

\begin{figure}[tbp]
	\centerline{\includegraphics[width=\linewidth]{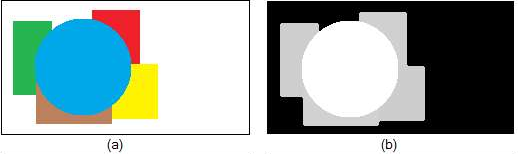}}
	\caption{An illustration for convex hull overlap (CHO) feature. (a) is an input image. The blue circle occupies some ares of the other objects. (b) Saliency map generated by CHO feature. The object's brightness indicates its saliency. The blue circle is the salient object in this image.}
	\label{illustration}
\end{figure}

\begin{figure}[tbp]
	\begin{subfigure}[t]{0.48\linewidth}
		\centering
		\includegraphics[height=2.8cm]{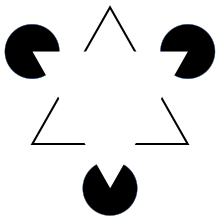}
		\caption{}
	\end{subfigure}%
	~
	\begin{subfigure}[t]{0.48\linewidth}
		\centering
		\includegraphics[height=2.8cm]{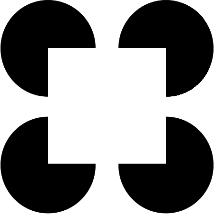}
		\caption{}
	\end{subfigure}
	\caption{Kanizsa triangle, after the Yugoslav psychologist Gaetano Kanizsa. Most people see a white triangle on the top because our brain automatically complete the black parts. It illustrates completion and convex hull reflect the overlap relationships among objects.}
	\label{Kanizsa_triangle}
\end{figure}

\begin{figure*}[tbp]
	\centerline{\includegraphics[width=\linewidth]{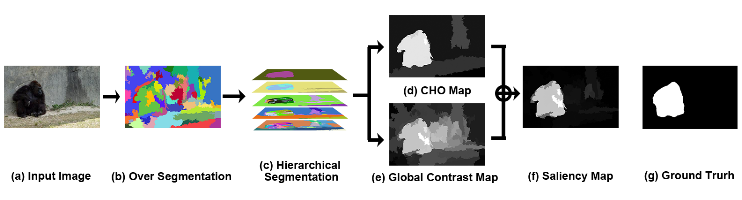}}
	\caption{The pipeline of our method. We use convex hull overlap feature and appearance contrast feature to detect salient objects.}
	\label{pipeline}
\end{figure*}

Besides that, both conventional saliency methods and CNN-based methods focus on appearance features, while they omit the importance of objects' positions. Figure \ref{illustration} is an illustration how the convex hull overlap feature capture the saliency by objects' positions. As the figure shown, since objects are all appearance different from the white background, it is difficult for those appearance driven methods to distinguish the most salient object. However, the blue one occupies other objects' contour convex hulls. It gives us an intuition that the blue object is on the top of other objects. People are attracted to the blue object due to this intuition. Our convex hull overlap feature is based on this intuition.

In cognition science, the research on salient object detection, especially the figure-ground analysis has been well studied since 1920s. Gestalt theory \cite{stevenson2012emergence} \cite{soegaard2010gestalt} believes that individuals perceive objects such as shapes, letters, pictures, as being whole when they are not complete. Specifically, when parts of a whole picture are missing, our perception fills in the visual gap. Convexity, one of the Gestalt laws, has been proved as a significant bottom-up feature to separate figure and background objects in perceptual organization. Figure \ref{Kanizsa_triangle} shows an example that the white part seems at the top of the black parts. Sterman \cite{sterman1994learning} explains this illusion that individuals' unconscious mental modeling tend to do meaningful completion automatically. Liu et al. \cite{liu1999role} argued for the importance of convexity in grouping and shown that convexity plays a role in perceptual grouping that cannot be accounted for by existing models based on good continuation.  The completion parts reflect the overlap relationships among objects. The convexity rule suggests that the object which occupies other objects' convex hulls tends to be a figure.

Based on the above observations, we establish a novel bottom-up feature named convex hull overlap (CHO), and then propose an effective approach to detect salient objects using the combination of the CHO feature and the appearance contrast feature. The hierarchical segmentation model is firstly constructed by using the recursive Normalized Graph-Cut algorithm \cite{shi2000normalized} to express the visual perception for objects and scenes more precisely. The convex hull is defined as the smallest convex set that contains all the pixels of a region in the Euclidean plane, and then the value of the CHO feature can be computed as the figural confidence of objects. Finally, the new feature description with the CHO feature can be effectively utilized to separate figural and background objects. Our contributions are summarized as follows:
\begin{itemize}
	\item We propose a robust hand crafted CHO feature. It can be used without training.
	\item Previous work only analyzes with appearance features, while our CHO feature makes up the obvious gap between the spatial region covering and the region saliency.
	\item A hierarchical segmentation model based on Normalized Graph-Cut fits the splitting and merging processes in human visual perception.
\end{itemize}

\section{Related Work}

Salient object detection belongs to the active research field of visual attention modeling. A major distinction among visual attention models is whether they rely on bottom-up features, top-down features, or a combination of both.

In conventional methods, intensity and color contrast features are simple and widely used in salient object detection. Zhai and Shah \cite{zhai2006visual} used luminance information to seek interesting regions in images. Achanta et al. \cite{achanta2009frequency} defined the pixel saliency using the pixels’ color difference from the average image color. Cheng et al. \cite{cheng2015global} computed the saliency map by evaluating global color contrast differences among regions. Rahtu et al. \cite{rahtu2010segmenting} adopted illumination, color and motion information with a Conditional Random Field model. Spectral information was utilized in Hou et al. \cite{hou2007saliency}, in which they analyzed the log-spectrum of an input and proposed a fast method to construct the corresponding saliency map in the spatial domain. Lu et al. \cite{lu2011salient} exploited convexity and surroundedness features, both of figure-ground features, to detect salient objects, but their model was sensitive to the super-pixels boundaries. Jiang et al. \cite{jiang2011automatic} integrated both the bottom-up salient stimuli and the object-level shape prior. Zhou et al. \cite{zhou2015salient} integrated compactness and local contrast cues to detect salient regions, but it may fail for images that do not have much color variation. Kim et al. \cite{kim2016salient} composited an accurate saliency map by finding the optimal linear combination of color coefficients in the high-dimensional color space. Scharfenberger et al. \cite{scharfenberger2015structure} proposed a structure-guided statistical textural distinctiveness approach to salient region detection. Zou et al. \cite{zou2015unsupervised} presented a novel unsupervised algorithm using Markov Random Field, but their method took long time calculations. Based on convexity feature, Palou and Salembier \cite{ palou2013monocular} focused on detecting special points, such as T-junctions and highly convex contours, to infer the depth relationships between objects in the scene. Tong et al. \cite{tong2015salient} proposed a bootstrap learning algorithm for salient object detection in which both weak and strong models are exploited. Yang et al. \cite{yang2013saliency} ranked the similarity of the image elements (pixels or regions) with foreground cues or background cues via graph-based manifold ranking. The saliency of the image elements is defined based on their relevances to the given seeds or queries. Jiang et al. \cite{jiang2013salient} used the supervised learning approach to map the regional feature vector to a saliency score, and finally fused the saliency scores across multiple levels, yielding the saliency map.

In recent years, deep learning based methods use high level features to detect salient regions. Lee et al. \cite{lee2016deep} concatenated low level distance map and high level VGG-net features to evaluate saliency. Li and Yu \cite{li2015visual} aggregated multiple saliency maps computed for different levels of image segmentation to boost the performance. Li et al. \cite{li2016deepsaliency} set up a multi-task learning scheme for exploring the intrinsic correlations between saliency detection and semantic image segmentation. Wang et al. \cite{wang2015deep} presented a saliency detection algorithm by integrating both local estimation and global search. Zhao et al. \cite{zhao2015saliency} took both global context and local context into account and jointly modeled in a unified multi-context deep learning framework. 

From another perspective of neuroscience and psychology, there exist an amount of research work on finding different ways to explore more beneficial features to illustrate the obvious appearances for salient regions through the chosen effective features. Wagemans et al. \cite{wagemans2012century} concluded that the connectivity and rules of visual cortex allowed the illusory contours to be formed and the figure-ground segmentation to be performed by the autonomous processes. Kimchi and Peterson \cite{kimchi2008figure} claimed that the figure-ground segregation could occur before the focal attention. It’s worth noting that the convexity cue of Gestalt laws, as widely believed, is a strong feature to assign the figure-ground relation. Bertamini and Wagemans \cite{bertamini2013processing} provided the evidence that the visual system could extract the information about convexities and concavities along with contours in an image. They concluded that the convexity affected the figure-ground relation. Fowlkes et al. \cite{fowlkes2007local} found that the convexity indeed had the ability to discriminate which part was the foreground on natural images by ecological statistics. Liu et al. \cite{liu1999role} demonstrated that stronger perceptual grouping causes a tendency in subjects to perceive planar figural elements as coplanar. This tendency can be measured by the extent to which it impedes the ability of subjects to detect non-coplanarity using stereo cues.

Unfortunately, all these existing methods have not yet provided good solutions for the following three crucial issues: 

\begin{enumerate}
	\item How does the objects' overlap relationship effects saliency -- We propose the convex hull overlap feature using spatial overlap relationship to reflect objects' saliency. 
	\item What is the general feature that effecting saliency the most -- In most cases, the color contrast and spatial information are sufficient to evaluate salient regions. Nevertheless, in the complex background, the background regions may be labeled as salient regions because of the sharp color contrast. In the deep neural network based methods, it is hard to explain what the meaning of extracted features are. So, this hand-crafted feature shows a posibble explanation of individuals' attention model.
	\item Improving the hierarchical segmentations -- Most existing methods build the hierarchical segmentation model by varying the number of super-pixels using the segmentation algorithm in Felzenswalb et al. \cite{felzenszwalb2004efficient}. In reality, an occluder may separate two regions from the same object. However, this information is lost in the previous hierarchical model.  Thus it’s necessary to set up a new hierarchical model by merging such two regions that belong to the same object into a larger region. 
\end{enumerate}

\begin{figure*}[tbp]
    \centerline{\includegraphics[width=\linewidth]{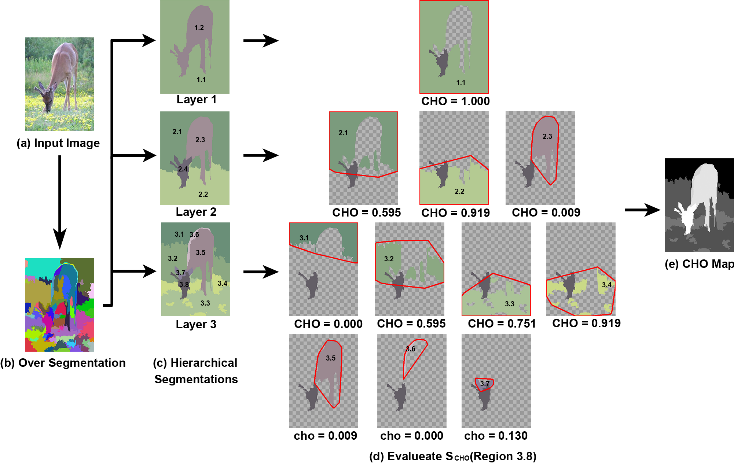}}
    \caption{Computation details of convex hull overlap feature in region $3.8$. As (d) shows, CHO value indicates how large area that the region $3.8$ occupies the other obejcts, which reflects the spatial overlap relationship. The front objects tend to be salient. In (e), brightness indicates saliency.}
    \label{details_CHO_maps}
\end{figure*}

\section{Hierarchical Segmentation}

This is the preprocessing process in our algorithm. Our goal is to build a $N$ hierarchical regions model from the input image. This will be used in saliency map calculation. Firstly, we over segment the input image into small regions. Then we merge two regions if their colors are similar. These two regions are not required to be connected.

\subsection{Over Segmentation}

Firstly, the input image is over segmented into small regions, denoted as $\{r_1, r_2, r_3, \cdots, r_M\}$, using Felzenszwalb et al.’s algorithm \cite{felzenszwalb2004efficient}, as shown in Figure \ref{pipeline}(b) and Figure \ref{details_CHO_maps}(b). These small regions will be merged to build a hierarchical regions model later.

\subsection{Hierarchical Regions Model}

A weighted graph $G=(V,E)$ is constructed by taking each region as a node and connecting each pair of regions by an edge. The weight on the edge, regarding to pixels' colors and spatial locations, reflects the likelihood that two regions belong to a larger region. The weight for the graph edge connecting two nodes of $r_i$ and $r_j$ is defined as:
\begin{equation}
w(r_i, r_j) = (|r_i|\cdot|r_j|)\cdot e^{\frac{D_c(r_i,r_j)}{-\sigma_c}}\cdot e^{\frac{D_s(r_i,r_j)}{-\sigma_s}}
\end{equation}
where $D_c(r_i,r_j)$ is the color difference between two regions, $\sigma_c$ is the color difference weight, $D_s(r_i, r_j)$ is the spatial distance between two regions, $\sigma_s$ is the spatial distance weight, and $|r_i|$ is the amount of pixels of region $r_i$.

Normalized Graph-Cut algorithm \cite{shi2000normalized} is recursively utilized to merge similar regions based on $G=(V,E)$ in order to construct a $N$ hierarchical regions model (in our experiments, we set $N=5$), as shown in Figure \ref{pipeline}(c) and Figure \ref{details_CHO_maps}(c). We define $L = \{l_1, l_2, l_3, \cdots , l_N\}$ as the layers of the hierarchical regions model, in which $l_1$ is the coarsest segmentation that contains 2 regions and $l_N$ is the finest one that contains at most $2^N$ regions.

\section{Saliency Map Computation}

We compute the saliency map for each regions in the hierarchical regions model. The computation integrates both convex hull overlap feature and appearance contrast (global contrast) feature.

\subsection{Convex Hull Overlap Saliency}

In our approach, a region consists of a set of pixel points. Convex hull $c$ is the smallest convex polygon, whose interior angles are less than $180$ degrees, which contains all the pixel points. We use Graham’s Scan algorithm to find the convex hull with the time complexity $O(nlogn)$, where $n$ is the number of pixel points. The gap between the region and its convex hull is filled with the pixels which belong to the other regions. We define the gap's area as the convex hull overlap feature. For example, a large gap's area means that this region is more likely to be a background region because its potential shape is occupied by other front regions.

We define the set of regions in layer $l_i$ as $\{r^i_1, r^i_2, r^i_3, \cdots, r^i_m\}$.
The convex hull $c^i_j$ is computed for each region $r^i_j$. 
Based on the related definition, the convex hull $c^i_j$ contains all the pixels of region $r^i_j$. 
Besides that, $c^i_j$ may also contain the pixels of other regions. 
Convex hull overlap feature of region $r^L_o$ is defined as follows:

\begin{equation}
	S_{CHO}(r^L_o) = \sum_{l=1}^{N}\frac{\sum_{j=1}^{m(l)}|c^l_j \cap r^L_o|}{m(l)}
\end{equation}
where $N$ is the number of layers, $m(l)$ is the number of regions in layer $l$, and $\cap$ is the intersection between two areas. If a region is a front region, the $S_{CHO}$ will be high since this region may occupy a lot of convex hulls of background regions. Figure \ref{details_CHO_maps}(d) shows the computational details of convex hull overlap feature. 

\subsection{Global Contrast Map}

Besides the convex hull overlap feature, people are highly sensitive to the appearance contrast in the visual signals. We calculate this feature $S_{GC}$ among all regions:
\begin{equation}
	S_{GC}(r_i) = w_c(r_i)\sum_{j}e^{\frac{D_s(r_i,r_j)}{-\sigma_s}}D_c(r_i,r_j)
\end{equation}
where $w_c(r_i)$ gives a low value if region $r_i$ is far from the center or a high value if the region is close to the center, $D_s(r_i, r_j)$ is the spatial distance between two regions, $\sigma_s$ is the spatial distance weight, $D_c(r_i,r_j)$ is the color difference between two regions. Figure \ref{pipeline}(e) shows the global contrast map.

\subsection{Saliency Map Generation}

In our algorithm, the saliency map takes into account both the convex hull overlap feature and the global contrast feature. For any pixel $p$ in the image, we define its saliency as follows.

\begin{equation}
    S(p) = \sum_{l=1}^{N} S_{CHO}(r_l) \cdot S_{GC}(r_l), p \in r_l
\end{equation}

Figure \ref{pipeline}(f) shows the pixel-level saliency map. The brightness indicates the confidence of saliency.

\section{Experiments}

\subsection{Dataset}

The performance of our approach is evaluated on the three publicly available datasets. Details of these datasets are as follows.

\begin{itemize}
	\item MSRA10K \cite{cheng2015global} contains $10,000$
	images with accurate human-marked labels for salient regions.
	\item ECSSD \cite{yan2013hierarchical} contains $1,000$ images and their multiple salient regions with complex backgrounds that make the detection task much more challenging.
	\item HKU-IS \cite{li2015visual} has 4447 images with high-quality pixel-wise	annotations.
\end{itemize}

\subsection{Evaluation Metrics}

The precision of a binary map is defined as the ratio of the intersection of salient area between estimation ($Sal$) and ground-truth ($GT$), to the salient area in this binary map. The recall value is the ratio of the intersection of salient area between estimation ($Sal$) and ground-truth ($GT$), to the salient area in the ground-truth map:

\begin{equation}
	precision = \frac{|Sal \cap GT|}{|Sal|}
\end{equation}

\begin{equation}
	recall = \frac{|Sal \cap GT|}{|GT|}
\end{equation}
where $|\cdot|$ denotes the area of the binary map in $1$ (or $255$).

The F-measure, denoted as $F_\beta$, is an overall performance indicator computed by the weighted harmonic of precision
and recall:

\begin{equation}
	F_\beta = \frac{(1+\beta^2)\cdot precision \cdot recall}{\beta^2 \cdot precission + recall}
\end{equation}
where $\beta^2$ is set to $0.3$ as suggested in \cite{achanta2009frequency} to emphasize the precision.

The mean absolute error (MAE) is another widely used evaluation metric which is the average per-pixel difference between the saliency map ($Sal$) and the ground-truth map ($GT$) \cite{hu2017deep}:

\begin{equation}
	MAE = \frac{1}{W \times H} \sum_{i=1}^H \sum_{j=1}^W |Sal(i,j) - GT(i,j)|
\end{equation}
where $W$ and $H$ are the width and height of the map respectively.

We still follow the official benchmark metrics of
precision and recall to measure the accuracy of our approach.
The precision-recall curve is exploited to reliably
compare how well various bottom-up features and saliency detection
methods highlight salient regions in images. Thus the
saliency value range of saliency map is normalized to $[0, 255]$,
and the threshold $T_f$ varies from $0$ to $255$ to binarize the
saliency map. The pixels are labeled as salient if their saliency
values are higher than $T_f$.

\subsection{Comparison with Conventional Methods}

\begin{table}[tbp]
	\caption{Comparison with Conventional Methods}
	\begin{center}
		\begin{tabular}{|c|c|c|c|c|c|c|c|}
			\hline
			\multirow{2}{*}{\centering\textbf{Dataset}}&\multirow{2}{0.4in}{\centering\textbf{Metric}}&\multicolumn{6}{|c|}{\textbf{Methods}} \\
			\cline{3-8}
			 & &\textbf{\textit{Ours}}&\textbf{\textit{RC}}&\textbf{\textit{HC}}&\textbf{\textit{LC}}&\textbf{\textit{SR}}&\textbf{\textit{FT}}\\
			\hline
			\multirow{3}{*}{\centering\textbf{ECSSD}}&Precision&\color{red}0.78&0.72&0.48&0.42&0.40&0.45\\
			\cline{2-8}
			&Recall&\color{red}0.63&0.62&0.46&0.38&0.45&0.45\\
			\cline{2-8}
			&$F_\beta$&\color{red}0.74&0.70&0.47&0.41&0.41&0.45\\
			\hline
			\multirow{3}{*}{\centering\textbf{MSRA}}&Precision&\color{red}0.89&0.83&0.72&0.62&0.51&0.70\\
			\cline{2-8}
			&Recall&\color{red}0.73&0.62&0.63&0.50&0.49&0.55\\
			\cline{2-8}
			&$F_\beta$&\color{red}0.85&0.77&0.70&0.59&0.50&0.65\\
			\hline
			\multicolumn{8}{p{0.95\columnwidth}}{Comparison between the proposed methods and conventional methods RC \cite{cheng2015global}, HC \cite{cheng2015global}, FT \cite{achanta2009frequency}, SR \cite{hou2007saliency} and LC \cite{zhai2006visual}. The best one is labeled in \color{red}red\color{black}.}
		\end{tabular}
		\label{tab1e_conventional}
	\end{center}
\end{table}

\begin{figure}[tbp]
	\centerline{\includegraphics[width=\linewidth]{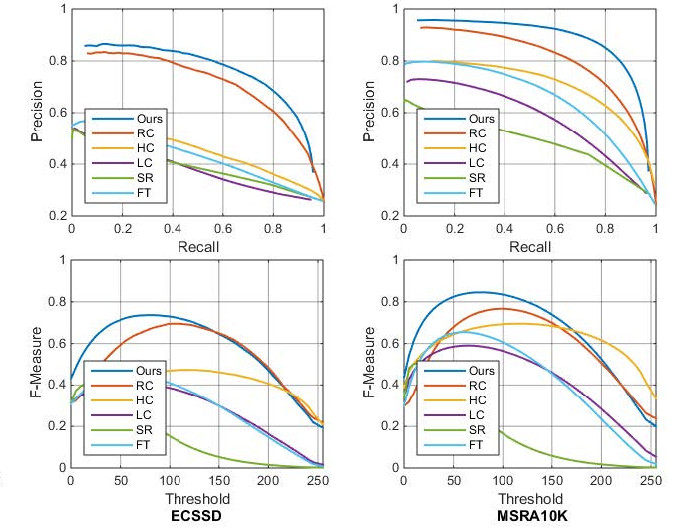}}
	\caption{The comparison results of the Precision-Recall curves and the F-Measure rates between the related conventional methods and ours
	on the dataset ECSSD \cite{yan2013hierarchical} and MSRA10K \cite{cheng2015global}.}
	\label{cmp_conventional}
\end{figure}

\begin{table*}[tbp]
	\caption{Comparison with Deep Learning Methods}
	\begin{center}
		\begin{tabular}{|c|c|c|c|c|c|c|c|c|c|c|c|c|c|}
			\hline
			\multirow{2}{*}{\centering\textbf{Dataset}}&\multirow{2}{0.4in}{\centering\textbf{Metric}}&\multicolumn{12}{|c|}{\textbf{Methods}} \\
			\cline{3-14}
			& &\textbf{\textit{Ours}}&\textbf{\textit{DLS}}&\textbf{\textit{ELD}}&\textbf{\textit{MDF}}&\textbf{\textit{MTDS}}&\textbf{\textit{LEGS}}&\textbf{\textit{MCDL}}&\textbf{\textit{BSCA}}&\textbf{\textit{GMR}}&\textbf{\textit{DRFI}}&\textbf{\textit{HC}}&\textbf{\textit{FT}}\\
			\hline
			\multirow{2}{*}{\centering\textbf{ECSSD}}&$F_\beta$&0.737&0.766&0.756&0.692&0.663&0.682&0.679&0.509&0.484&0.585&0.319&0.244\\
			\cline{2-14}
			&MAE&0.158&0.090&0.092&0.112&0.125&0.122&0.116&0.185&0.190&0.166&0.334&0.293\\
			\hline
			\multirow{2}{*}{\centering\textbf{HKU-IS}}&$F_\beta$&0.714&0.748&0.718&0.567&0.711&0.607&0.634&0.460&0.443&0.553&0.311&0.196\\
			\cline{2-14}
			&MAE&0.137&0.072&0.084&0.130&0.081&0.122&0.106&0.178&0.181&0.154&0.291&0.228\\
			\hline
			\multicolumn{14}{p{0.9\linewidth}}{Comparison between the proposed methods and deep learning based methods DLS \cite{hu2017deep}, ELD \cite{lee2016deep}, MDF \cite{li2015visual}, MTDS \cite{li2016deepsaliency}, LEGS \cite{wang2015deep}, MCDL \cite{zhao2015saliency}, BSCA \cite{tong2015salient}, GMR \cite{yang2013saliency}, DRFI \cite{jiang2013salient},  HC \cite{cheng2015global} and FT \cite{achanta2009frequency}.}
		\end{tabular}
		\label{tab1e_deep}
	\end{center}
\end{table*}

\begin{figure}[tbp]
	\centerline{\includegraphics[width=\linewidth]{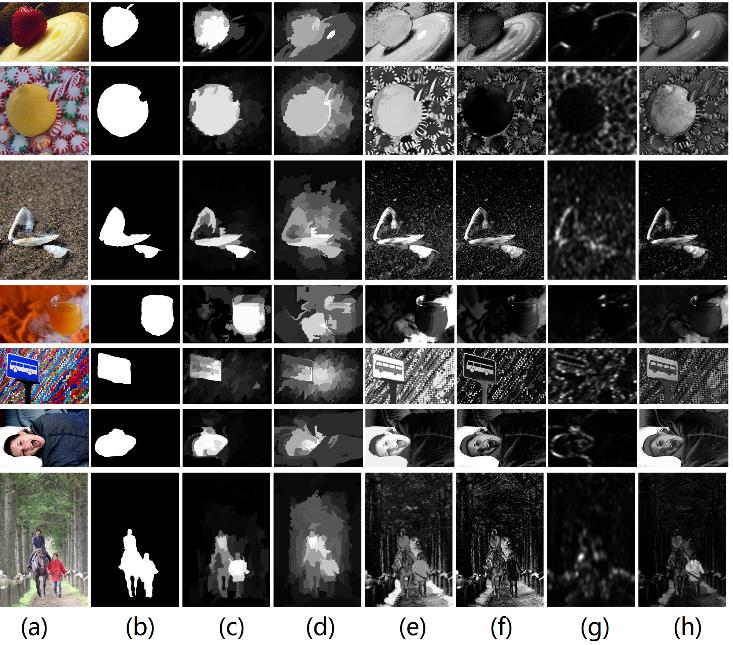}}
	\caption{Selected results between the related conventional methods and ours, in which (a) Input image; (b) Ground truth; (c) Ours; (d) RC\cite{cheng2015global}; (e) HC\cite{cheng2015global}; (f) LC\cite{zhai2006visual}; (g) SR\cite{hou2007saliency}; (f) FT\cite{achanta2009frequency}.}
	\label{result_conventional}
\end{figure}

To objectively evaluate our salient region detection pattern, we re-implement other conventional methods. The re-implement methods' performance are slightly weaker than their paper reported. Then we use both the fixed threshold and different threshold setting to compare our approach with the other related conventional methods. The precision-recall curves with the fixed threshold and the F-measure rates under different threshold settings are shown in Figure \ref{cmp_conventional}. Under the same recall rate, our salient region detection pattern exhibits the higher precision, which indicates our approach is robust and effective. When the threshold $T_f$ is low, our approach can achieve the best F-measure. The reason is that our method can separate the salient regions and non-salient regions robustly in the boundary.

Furthermore, Table \ref{tab1e_conventional} shows the quantitative comparison analysis between the related conventional methods and ours with the best F-measure rate. These results also reveal that our approach can actually achieve more competitive performance compared to the other existing methods. This is because the color information does not work well in the complex background, while the geometrical and covering information can be still effective to distinguish the figural and background regions. It also indicates that the in-depth salient object analysis by using our approach with the convex hull overlap feature is necessary and efficient to improve the salient object judgment.

\begin{figure*}[tbp]
	\centerline{\includegraphics[width=0.87\linewidth]{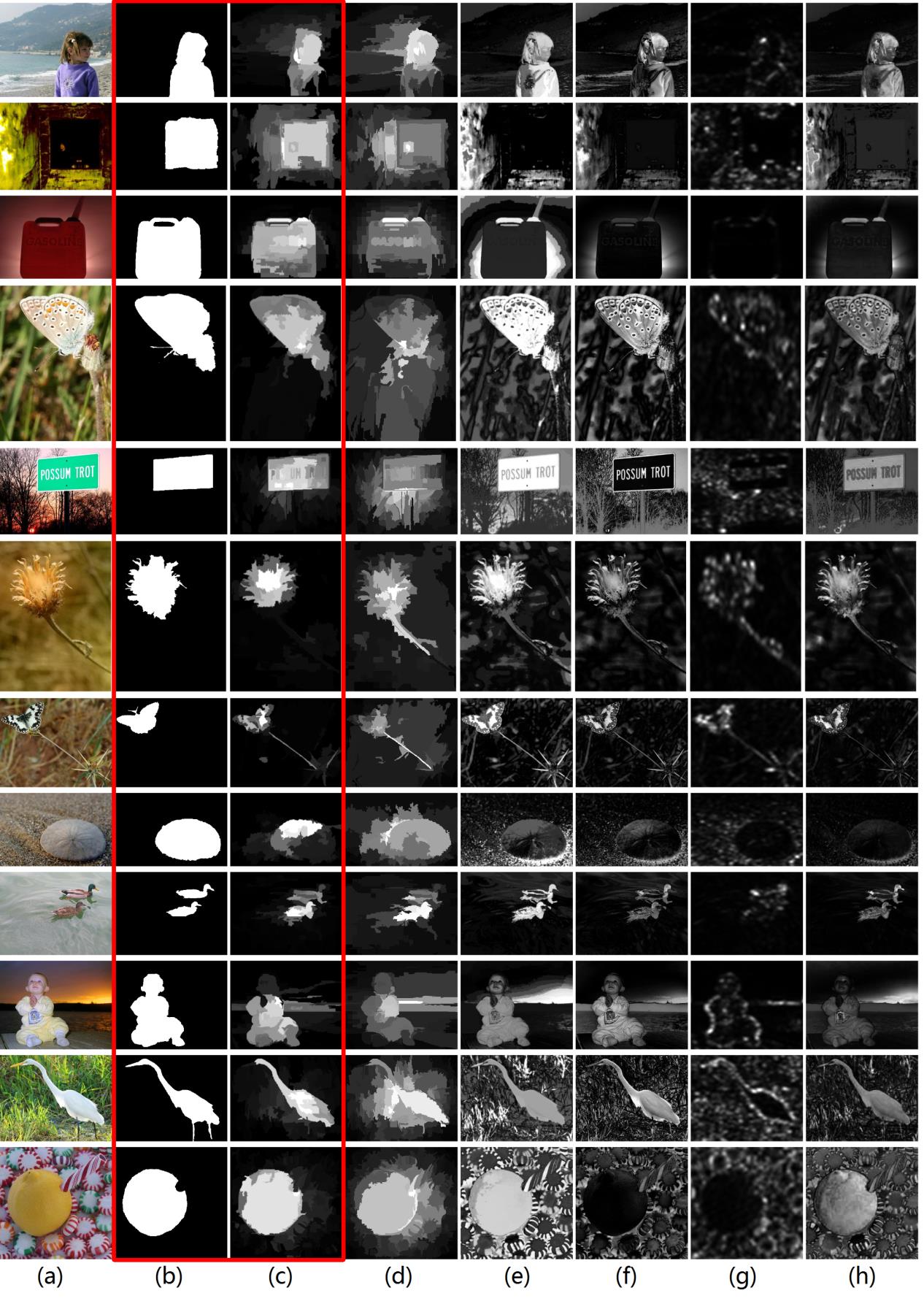}}
	\caption{More results between the related conventional methods and ours, in which (a) Input image; (b) Ground truth; (c) Ours; (d) RC\cite{cheng2015global}; (e) HC\cite{cheng2015global}; (f) LC\cite{zhai2006visual}; (g) SR\cite{hou2007saliency}; (f) FT\cite{achanta2009frequency}.}
	\label{result_conventional_more_0}
\end{figure*}

\subsection{Comparison with Deep Learning Methods}

We also compare our methods with the recent state-of-the-art deep learning methods on ECCSD and HKU-IS. We use the comparison data from \cite{hu2017deep} as Table \ref{tab1e_deep}.

Compared the hand-crafted feature results and the deep learning based results, it can easily find out that deep learning methods have better performance than the conventional methods. The reasons is that deep learning methods extract feature from the learning dataset, while the hand-crafted feature is a kind of prio knowledge based feature. However, our proposed method's performance is closed to the state-of-the-art deep learning based methods. It suggests that our proposed method capture the key information about the description of saliency. Besides that, we can explain little physical meaning of the deep learning based feature, but our convex hull overlap feature make up the gap between the spatial overlap relationship and object's saliency. It shows a possible explanation of the individuals' attention model.

\section{Conclusion}

A new framework is introduced in this paper to support more precise salient region detection. The convex hull overlap feature based on Gestalt laws is especially introduced to make up the gap between the object overlap relationship and saliency. It also achieves an optimal representation for images. Different from machine learning, our scheme using hand-crafted feature has advantages in the results and computational consuming. Our future work will focus on setting up a specialized learning mechanism to implement the deep-level analysis and mining for in-depth features.

\bibliographystyle{IEEEtran}
\bibliography{refs}

\end{document}